\documentclass[manuscript,screen]{acmart}

\AtBeginDocument{%
  \providecommand\BibTeX{{%
    \normalfont B\kern-0.5em{\scshape i\kern-0.25em b}\kern-0.8em\TeX}}}

\copyrightyear{2021}
\acmYear{2021}
\setcopyright{rightsretained}
\acmConference[RecSys '21]{Fifteenth ACM Conference on Recommender Systems}{September 27-October 1, 2021}{Amsterdam, Netherlands}
\acmBooktitle{Fifteenth ACM Conference on Recommender Systems (RecSys '21), September 27-October 1, 2021, Amsterdam, Netherlands}\acmDOI{10.1145/3460231.3474609}
\acmISBN{978-1-4503-8458-2/21/09}

\begin{document}

\title{Learning to Match Job Candidates Using Multilingual Bi-Encoder BERT}

\author{Dor Lavi}
\email{dor.lavi@randstadgroep.nl}
\affiliation{%
  \institution{Randstad Groep Nederland}
  \city{Diemen}
  \country{The Netherlands}
}



\maketitle

\section*{Abstract}

Randstad is the global leader in the HR services industry. We support people and organizations in realizing their true potential by combining the power of today’s technology with our passion for people. In 2020, we helped more than two million candidates find a meaningful job with our 236,100 clients. Randstad is active in 38 markets around the world and has top-three positions in almost half of these. In 2020, Randstad had on average 34,680 corporate employees and generated revenue of € 20.7 billion.

Each day, at Randstad, we employ industry-scale recommender systems to recommend thousands of candidates to our clients, and the other way around; vacancies to job seekers. Our ``Talent Recommender'' recommender system is based on a heterogeneous collection of input data: CVs, vacancy texts (job descriptions) and structured data (e.g., the location of a candidate or vacancy). The goal of the system is to recommend the best candidates (talents) to each open vacancy. 

CVs are user-generated PDF files. It goes without saying that parsing those files to plain text can be a challenge in itself and therefore out of scope for this talk. On the other hand, vacancies are usually structured formatted text. We should be aware that due to the difference in structure and preprocessing steps, that the input to the subsequent steps is inevitably noisy. 

Most NLP research in text similarity is based on the assumption that 2 pieces of information are the same but written differently~\cite{10.1145/3440755}. Like two artists that paint the same landscape, but each with its own style. However, in our case the 2 documents complement one another like pieces in a puzzle, together they create the bigger picture, rather than 2 similar paintings.

Some of our biggest challenges with the ``Talent Recommender'' stem from dealing with the diverse nature of our textual sources of data: vacancies and CVs. While both capture similar information, they are inherently different in many ways. 

First, the information in CVs and vacancies are similar, but there exists a vocabulary gap, where grammar and context differ. For example, where \emph{``I have 10 years of experience as an instructor''} in a CV shares no word overlap with \emph{``We are looking for a talented tutor''} in a  vacancy, both cases express similar information regarding \emph{``experience in the field of education,''} we need to overcome the synonyms gap \emph{``instructor''} and \emph{``tutor.''} In addition, the sentence structure is completely different, CVs are typically written in ``storytelling mode'' \emph{``I have\ldots,''} while the vacancy is in ``exploration mode'' \emph{``we are looking\ldots''} 

The second challenge is multilinguality. Since we are a multinational company that operates all across the globe, developing a model per language is not scalable in our case. We ultimately would like one maintainable model that supports as many languages as possible.

Our last challenge is cross language similarity~\cite{10.1613/jair.1.11640}. In some of the countries we operate, there is a high percentage of job seekers that are not native to that country. For example, many of the job descriptions in the Netherlands are in Dutch, however around 10\% of the CVs are in English. Classic text models, like TF-IDF and Word2vec, capture information within one language, but hardly connect between languages. Simply put, even if trained on multiple languages each language will have its own cluster in space. So \emph{``logistics''} in English and \emph{``logistiek''} in Dutch are embedded in a completely different point in space, even though the meaning is the same. Furthermore, we know that the language of CV correlates with nationality and therefore can be a proxy discriminator. Due to the impact of these systems and the risks of unintended algorithmic bias and discrimination, HR is marked as a high risk domain in the recently published EC Artificial Intelligence Act~\cite{european2021proposal}. To avoid discriminating against nationality we would like to recommend a candidate to the vacancy no matter which language the CV is written in. That is of course only if language is not a requirement for that vacancy.

In this talk, we will show how we used our internal history of candidate placements to generate labeled CV-vacancy pairs dataset. Afterwards we fine-tune a multilingual BERT with bi encoder structure~\cite{reimers-gurevych-2019-sentence} over this dataset, by adding a cosine similarity log loss layer. We will explain how using the mentioned structure helps us overcome most of the challenges described above, and how it enables us to build a maintainable and scalable pipeline to match CVs and vacancies. In addition, we show how we gain a better semantic understanding, and learn to bridge the vocabulary gap. Finally, we highlight how multilingual transformers help us handle cross language barrier and might reduce discrimination.

\section*{Speaker Bio}
\textbf{Dor Lavi} is a senior data scientist at Randstad. Dor has experience in both academia and industry environments. He worked as a researcher in Tel Aviv University. His research focused on the impact of Israeli regulation on the real estate market in Israel. He worked as a data scientist in a variety of technology companies, some of them are tech giants such as Booking.com and others are smaller scale ups like Similarweb. At Randstad, Dor is responsible for the core recommendation system for that match between job seekers and vacancies.

\bibliographystyle{ACM-Reference-Format}
\bibliography{bibliography}

\end{document}